\documentclass[10pt,journal,transmag,compsoc]{IEEEtran}
\usepackage{array}
\usepackage{subfig}
\usepackage{eurosym}
\pdfoutput=1
\newenvironment{conditions}
  {\par\vspace{\abovedisplayskip}\noindent\begin{tabular}{>{$}l<{$} @{${}={}$} l}}
  {\end{tabular}\par\vspace{\belowdisplayskip}}
\ifCLASSOPTIONcompsoc
  \usepackage[nocompress]{cite}
\else
  \usepackage{cite}
\fi
\ifCLASSINFOpdf
  \usepackage[pdftex]{graphicx}
  \graphicspath{{../figs/}}
  \usepackage{epstopdf}
  \DeclareGraphicsExtensions{.pdf,.jpeg,.png,.eps}
\else
  \usepackage[dvips]{graphicx}
    \usepackage{epstopdf}

  \graphicspath{{../figs/}}
  \DeclareGraphicsExtensions{.eps}
\fi
\usepackage{amsmath}

\usepackage{amsmath}
\usepackage{csvsimple}

\usepackage{fixltx2e}

\usepackage{stfloats}
\usepackage{url}
\begin{document}
%

\title{Improved Reinforcement Learning with Curriculum}
%
%
%
%

\author{\IEEEauthorblockN
{
Joseph~West\IEEEauthorrefmark{1},
Frederic~Maire\IEEEauthorrefmark{1},~\IEEEmembership{Member,~IEEE,}
~Cameron~Browne\IEEEauthorrefmark{2} and
~Simon Denman\IEEEauthorrefmark{1},~\IEEEmembership{Member,~IEEE}
}
        
\IEEEauthorblockA{\IEEEauthorrefmark{1}School of Electrical Engineering and Computer Science, Queensland University of Technology, Brisbane, QLD, 4500, Australia}

\IEEEauthorblockA{\IEEEauthorrefmark{2}Department of Data Science and Knowledge Engineering, Maastricht University, Maastricht, 6211, Netherlands}%

\thanks{Manuscript received xxxx, 2019; 
\emph{Corresponding Author: jd.west@hdr.qut.edu.au}
}
}
\IEEEdisplaynontitleabstractindextext
\IEEEpeerreviewmaketitle
%
%

\markboth{IEEE TRANSACTIONS ON GAMES,~Vol.~XX, No.~X, March~2019}%
{Improved Reinforcement Learning with Curriculum}
\IEEEtitleabstractindextext{%
\begin{abstract}
Humans tend to learn complex abstract concepts faster if examples are presented in a structured manner. For instance, when learning how to play a board game, usually one of the first concepts learned is how the game ends, i.e. the actions that lead to a terminal state (win, lose or draw). The advantage of learning end-games first is that once the actions which lead to a terminal state are understood, it becomes possible to incrementally learn the consequences of actions that are further away from a terminal state - we call this an \emph{end-game-first} curriculum. Currently the state-of-the-art machine learning player for general board games, \emph{AlphaZero} by \emph{Google DeepMind}, does not employ a structured training curriculum; instead learning from the entire game at all times. By employing an \emph{end-game-first} training curriculum to train an \emph{AlphaZero} inspired player,
we empirically show that the rate of learning of an artificial player can be improved when compared to a player not using a training curriculum.
\end{abstract}

}

\maketitle

\IEEEdisplaynontitleabstractindextext

%
\IEEEpeerreviewmaketitle
\setcounter{secnumdepth}{5}

\ifCLASSOPTIONcompsoc
\IEEEraisesectionheading{\section{Introduction}\label{sec:introduction}}
\else
\section{Introduction}
\label{sec:introduction}
\fi
\IEEEPARstart{A}n artificial game playing agent can utilise either a knowledge-based method or a brute-force method to determine which move to make when playing a game~\cite{van_den_herik_games_2002}. 
Traditionally, brute-force methods like uninformed tree-search are effective for games with low state-space complexity whilst pure knowledge-based methods like direct-coding or neural-networks are generally best for games with a low decision complexity~\cite{van_den_herik_games_2002}. 
The current state-of-the-art game playing agent, \emph{Google DeepMind's AlphaZero}~\cite{silver_mastering_2017} uses a combination of brute-force and knowledge-based methods by using a neural-network to guide a tree-search to decide its moves when playing a game. Whilst \emph{AlphaZero} has demonstrated superhuman performance on a number of different games, we address a weakness in its method and propose an approach which results in an improvement in the time the agent takes to learn.  

\emph{AlphaZero} generates its own training examples as part of its learning loop through \emph{self-play} and the generated examples continually improve as the network learns via \emph{reinforcement learning}~\cite{sondik_optimal_1978,sutton_learning_1988,sutton_temporal-difference_2015}. 
Neural-networks learn via \emph{reinforcement learning} by minimising the difference between the network's prediction for a state at one time and a \emph{future prediction} for the same state at some later time, called \emph{temporal difference (TD) learning} \cite{sutton_learning_1988}.
The \emph{future prediction} can be approximated by conducting a tree-search to \emph{look-ahead}, this variation of TD learning is called \emph{TD-Leaf} \cite{baxter_tdleaflambda_1999} and underpins \emph{AlphaZero's} combined tree-search/neural-network design, that is; \emph{AlphaZero} uses the difference between the neural-network's prediction and the outcome of a tree-search to train its network. 


\emph{AlphaZero} does not use its neural-network to directly make move decisions, instead it is used to identify the \emph{most promising} actions for the search to explore and to also estimate the values of states which are non-terminal. The benefit of this design is that as the neural-network improves the tree-search also improves, however, a characteristic of \emph{reinforcement learning} is that the network's predictions are initially inaccurate. 
\emph{AlphaZero's} method is effective, however many of the early training examples are of inherently poor quality as they are generated using an \emph{inadequately trained}\footnote{A network can be \emph{inadequately trained} either due to poor training or insufficient training.} neural-network.  

The neural-network used in \emph{AlphaZero} has the net effect of identifying an initial set of \emph{most promising} paths for the tree-search to explore. As the search is conducted, if the initial \emph{most promising} paths are poorly selected by the neural-network they can be overridden by a sufficiently deep search, resulting in a good move decision despite the poor initial \emph{most promising} set. A weakness with this approach arises when:
\begin{itemize}
\item the network is inadequately trained, typically during the early stages of training; \textbf{and} 
\item the tree search is less likely to discover terminal states, typically in the early stages of a game. 
\end{itemize}
If the network is inadequately trained, then the set of moves that the network selects, which are intended to be promising moves, will likely be random or at best poor.  
If the tree-search does not find sufficient terminal states then the expansion of the tree is primarily determined by the neural-network, meaning that there may not be enough actual game-environment rewards for the tree-search to correct any poor network predictions. If both of these situations occur together then we say that the resulting decision is \emph{uninformed}, which results in an \emph{uninformed} training example that has little or no information about the game that we are trying to learn. 
In this paper we demonstrate how employing an \emph{end-game-first} training curriculum to exclude expectedly \emph{uninformed} training experiences results in an improved rate of learning for a combined tree-search/neural-network game playing agent. 
\subsection{Outline of Experiment}
The effect of an \emph{end-game-first} training curriculum can be achieved by discarding a fraction of the early-game experiences in each self-play game, with the fraction dependent on the number of training epochs which have occurred. 
By selectively discarding the early-game experiences, we hypothesise that the result is a net improvement in the quality of experiences used for training which leads to the observed improvement in the rate of learning of the player. 
We demonstrate the effectiveness of the \emph{end-game-first} training curriculum on two games: \emph{modified Racing-Kings} and \emph{Reversi}. 
During training, we compare the win-ratio's of two \emph{AlphaZero} inspired game playing agents: a \emph{baseline player} without any modification, and a player using an \emph{end-game-first} curriculum \emph{(curriculum player)}; against a fixed \emph{reference opponent}. 

We find that by using only the late-game data during the early training epochs, and then expanding to include earlier game experiences as the training progresses, the \emph{curriculum player} learns faster compared to the \emph{baseline player}. Whilst we empirically demonstrate that this method improves the player's win-ratio over the early stages of training, the curricula used in this paper were chosen semi-arbitrarily, and as such we do not claim that the implemented curricula are optimal. We do however show that an \emph{end-game-first} curriculum learning approach improves the training of a combined tree-search/neural-network game-playing reinforcement learning agent like \emph{AlphaZero}.

The structure of the remainder of this paper is as follows. We review the related work in Section \ref{sec:related}; then outline the design of our system in Section \ref{sec:system}; present our evaluation in Section \ref{sec:eval}; and finally we conclude the paper in Section \ref{sec:conc}. For ease of understanding we explain in this paper how our proposed approach enhances \emph{AlphaZero}'s method and for the sake of conciseness only revisit key concepts from \cite{silver_mastering_2017} where they are needed to the explain our contribution.      

\subsection{Related Work}
\label{sec:related}
\subsubsection{AlphaZero's Evolution}
\emph{Google Deepmind's AlphaGo} surprised many researchers by defeating a professional human player at the game of Go in 2016~\cite{silver_mastering_2016}, however it was highly customised to the game of Go. In 2017 \emph{AlphaGo~Zero} was released, superseding \emph{AlphaGo} with a generic algorithm with no game-specific customisation or hand-crafted features, although only tested on Go\cite{silver_mastering_2017_1}. The \emph{AlphaGo~Zero} method was quickly validated as being generic with the release of \emph{AlphaZero}, an adaption of \emph{AlphaGo Zero} for the games of Chess and Shogi~\cite{silver_mastering_2017}. Both \emph{AlphaGo~Zero} and \emph{AlphaGo} had a three stage training pipeline; self-play, optimization and evaluation as shown in Figure \ref{fig:azloop} and explained in Section \ref{agzinspired}. \emph{AlphaZero} differs primarily from \emph{AlphaGo~Zero} in that no evaluation step is conducted as explained in Section \ref{training}. While \emph{AlphaGo} utilises a neural-network to bias a monte-carlo-tree-search (MCTS) expansion~\cite{kocsis_bandit_2006, browne_survey_2012}, \emph{AlphaGo~Zero} and \emph{AlphaZero} completely exclude conducting monte-carlo rollouts during the tree-search, instead obtaining value estimates from the neural-network. This MCTS process is explained in Section \ref{MCTS}.       


\subsubsection{Using a Curriculum for Machine Learning}

Two predominant approaches are outlined in the literature for curriculum learning in relation to machine learning: \emph{reward shaping} and \emph{incrementally increasing problem complexity}~\cite{elman_learning_1993}. 

\emph{Reward shaping} is where the reward is adjusted to encourage behaviour in the direction of the goal-state, effectively providing sub-goals which lead to the final goal. For example, an agent with an objective of moving to a particular location may be rewarded for the simpler task of progressing to a point in the direction of the final goal, but closer to the point of origin. When the sub-goal can be achieved, the reward is adjusted to encourage progression closer to the target position~\cite{asoh_socially_1997}. Reward shaping has been used successfully to train an agent to control the flight of an autonomous helicopter in conducting highly non-linear manoeuvres~\cite{ng_policy_1999, ng_autonomous_2004, ng_autonomous_2006}.

Another approach to curriculum learning for a neural-network is to incrementally increase the problem complexity as the neural-network learns~\cite{lee_learning_2011}. This method has been used to train a network to classify \emph{geometric shapes} by using a two step training process. In this approach, a simpler training set was used to initially train the network, before further training the network with the complete dataset which contained additional classes. This approach resulted in an improved rate of learning~\cite{bengio_curriculum_2009} for a simple neural-network classifier. Both \emph{reward shaping} and \emph{incrementally increasing problem complexity} rely on prior knowledge of the problem and some level of human customisation~\cite{jiang_self-paced_nodate, kumar_self_2010}. 

Florensa \textit{et al.} \cite{florensa_reverse_nodate} demonstrated how an agent can learn by reversing the problem-space, by positioning an agent at the goal-state and then exploring progressively further away from the goal. Their method, \emph{reverse curriculum}, sets the agent's starting position to the goal-state and noisy actions are taken to obtain other valid states in close proximity to the goal. As the agent learns, the actions result in the agent moving further and further from the goal. Their method was demonstrated to improve the time taken to train the agent in a number of single-goal reinforcement learning problems including a robot navigating a maze to a particular location and a robot arm conducting a single task such as placing a ring on a peg. The set of problems which the \emph{reverse curriculum} is suited to is constrained due to the requirement of a known goal state, and that the goal states themselves be non-terminal; i.e. on reaching the goal-state legal actions still exist which move the agent away from the goal. Having a known goal state permits the \emph{reverse curriculum} to be particularly useful if the problem's goal state is unlikely to be discovered through random actions. The weakness with the \emph{reverse curriculum} is that if the problem has a number of distinct goal-states, then focusing on training near a single known goal would result in the agent being over-fitted to the selected known goal at the exclusion of all others. The \emph{reverse curriculum} is not suitable for games, as games have multiple terminal states which are not known in advance, and once a terminal state is reached there are no further legal actions that can be taken.

We define an \emph{end-game-first} curriculum as one where the initial focus is to learn the consequences of actions near a terminal state/goal-state and then progressively learn from experiences that are further and further from the terminal state. We consider the \emph{reverse curriculum} to be a special case of the \emph{end-game-first} curriculum due to the additional requirements outlined above. It is not a requirement of the \emph{end-game-first} curriculum that the agent commence near a terminal/goal state, or that one is even known; but in the course of exploration when a terminal/goal state is discovered the distance to any visited non-goal states can be calculated and a decision can be made as to whether those states will be used for training the agent depending on their distance. 

The \emph{end-game-first} curriculum differs from \emph{incrementally increasing the problem complexity} in that the consequences of actions leading to a terminal state may in fact be more complex to learn than the earlier transitions. The \emph{end-game-first} curriculum does however temporarily reduce the size of the problem space by initially training the network on a smaller subset of the overall problem space.
The advantage of an \emph{end-game-first} curriculum is that it doesn't rely on any prior knowledge of the problem - including requiring a known terminal state. By first focusing near a terminal state, the agent is trained to recognise the features and actions which give rise to environmental rewards (terminal states) then progressively learns how to behave further and further from these states.  
We demonstrate in this paper that using an \emph{end-game-first} curriculum for training a combined tree-search/neural-network game playing agent can improve the rate at which the agent learns.  

\section{System Design} 
\label{sec:system}

The system consists of two core modules: the game environment and the players. Both the environment and the players utilise a standard framework, regardless of the game or the type of player. The system is designed in such a way as to reduce the variability between compared players. The agent's performance during both training and gameplay is traded off in favour of reducing the variability of the experiments. For example, the training pipeline used for the neural-network is conducted sequentially, however better performance could be obtained by conducting the training in parallel. Likewise, when two players are compared they are both trained on the same system simultaneously to further reduce the impact of any variance in processor load. The result of this design decision is that the complexity of the agent, the difficulty of the games and the quality of the opponents are constrained to permit the simultaneous training of two AI agent's, using sequential processes, within a period of time which is reasonable yet is still sufficiently complex to demonstrate the effectiveness of the presented method. Our focus is not on the absolute performance of the agent with respect to any particular game, but instead the comparative improvement of using an \emph{end-game-first} training curriculum.               

\subsection{Game Environment}
\label{sec:gameEnv}

The game environment encodes the rules and definitions of the game as well as maintains the progress of the game, permitting players to make moves and updating the game state accordingly. The players query the game environment to inform their decisions. A game has a set of $m$ possible actions,  {${A}=\{a_{1},..,a_{m}\}$}. For any game state, the environment provides:
\begin{itemize}
\item a tensor of sufficient size to represent the current state {${s}$}; 
\item a bit array {${b}{(s)}$} of length $m$ with bits representing the legal actions set to 1;
\item a vector representing the set of legal actions {${L}{(s)}\subset {A}$} in a format accepted by the environment's \emph{move} function - internally mapped to {${b}{(s)}$}; NB that {${A}$} is used to denote the set of all actions while $a$ is used to denote a single action; and
\item a scalar $w$ indicating if the game: is ongoing, a draw, player 1 wins, or player 2 wins. 
\end{itemize}

\subsubsection{Games}
\label{sec:games}
We evaluate the effectiveness of using an \emph{end-game-first} curriculum on two games; a modified version of Racing Kings \cite{noauthor_racing_nodate} and Reversi \cite{landau_othello}. The games selected have fundamental differences in how a player makes their move. Reversi is a game which the board starts nearly empty and as the game progresses, tiles are placed on the board filling the board up until the game ends; tiles are not relocated once they are placed. Games like Tic-Tac-Toe, Connect Four, Hex and Go have the same movement mechanics as Reversi and with the exception of the occasional piece removal in Go, these games also fill the board as the game progresses. Racing Kings was chosen as the representative of the class of games that have \emph{piece mobility}, i.e. where a piece is moved by picking it up from  one cell and placing it at another. Like \emph{Racing Kings}, often games with \emph{piece mobility} also have the characteristic of piece removal through capture. Games with similar movement mechanics as Racing Kings includes the many Chess variants, Shogi and Nim.

\begin{figure}
\begin{minipage}[b]{.99\columnwidth}
    \centering
    \subfloat[Racing Kings]{\includegraphics[width=.5\textwidth]{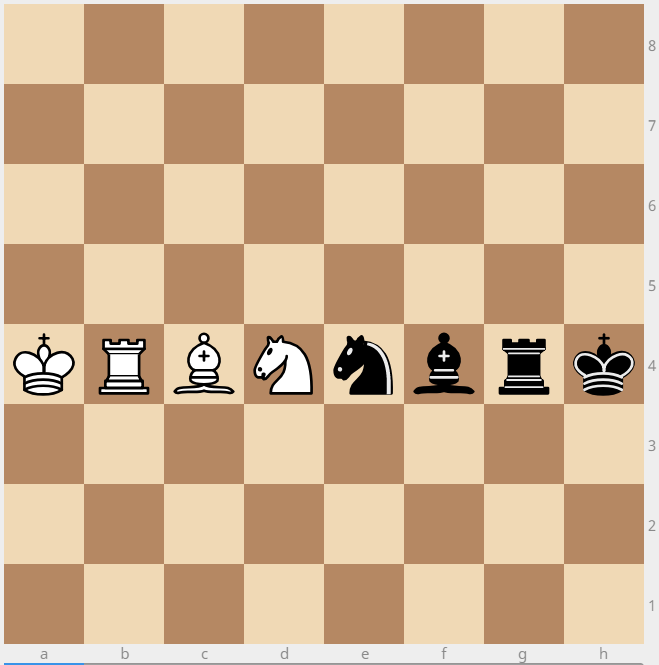}}%
    \subfloat[Reversi]{\includegraphics[width=.5\textwidth]{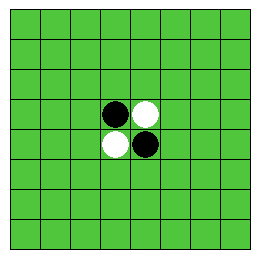}}%
    \caption{Starting board positions for Modified Racing Kings and Reversi. The aim of Racing Kings is to move your King to the final rank (top row), while the aim of Reversi is to have more pieces on the board when neither player can no longer move.}%
    \label{fig:rkStart}

   \end{minipage}

\end{figure}

\paragraph{Racing Kings} Racing Kings is a game played on a Chess board with Chess pieces. Pieces are placed on a single row at one end of the board with the aim being to be the first player to move their King to the other end of the board. Checkmate is not permitted, and neither is castling, however pieces can be captured and removed from the board. 
We modify the full Racing Kings game by using less pieces and adjusting the starting position of the pieces by placing them in the middle of the board, as shown in Figure \ref{fig:rkStart}, instead of on the first rank. As the Racing Kings game library is part of a suite of Chess variants \cite{fiekas_pure_2019} we maintain the environment as it would be for Chess. A state ${s}$ is represented as a tensor of size $8\times8\times12$; the width and height dimensions representing cells on the board and the $12$ planes representing each of the player's pieces; King, Queen, Rook, Bishop, Knight and Pawn. As the pieces are moved from one cell to another the total number of actions includes all possible pick and place options, $64\times64=4096$, excluding any additional actions such as promotion of pawns. There are $44$ possible \emph{from/to} pawn promotion movements in Chess but this also needs to be multiplied by the number of pieces which a pawn can be promoted to. We only consider promotion to a Knight or a Queen giving $88$ possible promotion actions. The total number of actions for the Racing Kings game environment is $m=4184$. 

\paragraph{Reversi} Reversi ``is a strategic boardgame which involves play by two parties on an eight-by-eight square grid with pieces that have two distinct sides ... a light and a dark face. There are 64 identical pieces called `disks'. The basic rule of Reversi, if there are player’s discs between opponent’s discs, then the discs that belong to the player become the opponent’s discs''\cite{gunawan_evolutionary_2012}. The winner is the player with the most tiles when both players have no further moves. For the Reversi game environment a state ${s}$ is represented as a tensor of size $8\times8\times2$; the width and height dimensions representing cells on the board and each plane representing the location of each players' pieces, with 64 possible different actions $m=64$. Our Reversi environment is based on \cite{morishita_reversi_2019}.    


\subsection{Player Architecture}
Two \emph{Artificial Intelligence} (AI) players: a \emph{baseline player} without any modifications, and a \emph{curriculum player} using a specified \emph{end-game-first} training curriculum, are tested in a contest against a game specific \emph{reference opponent}. 

\subsubsection{Artificial Intelligence Players} 

The architecture of both players is a combined neural-network/MCTS reinforcement learning system which chooses an action $a$ from the legal moves {${L}{(s)}$} for any given state {${s}$}. The neural-network with parameters {${\theta}$} is trained by self-play reinforcement learning. As the players are inspired by \emph{AlphaZero}\cite{silver_mastering_2017}, only the components relevant to this paper will be covered in this section. 

\vspace{\baselineskip}\noindent\paragraph{Neural-Network}

\begin{figure*}
   \begin{minipage}[b]{1\textwidth}
   \centering
   \includegraphics[width=\textwidth]{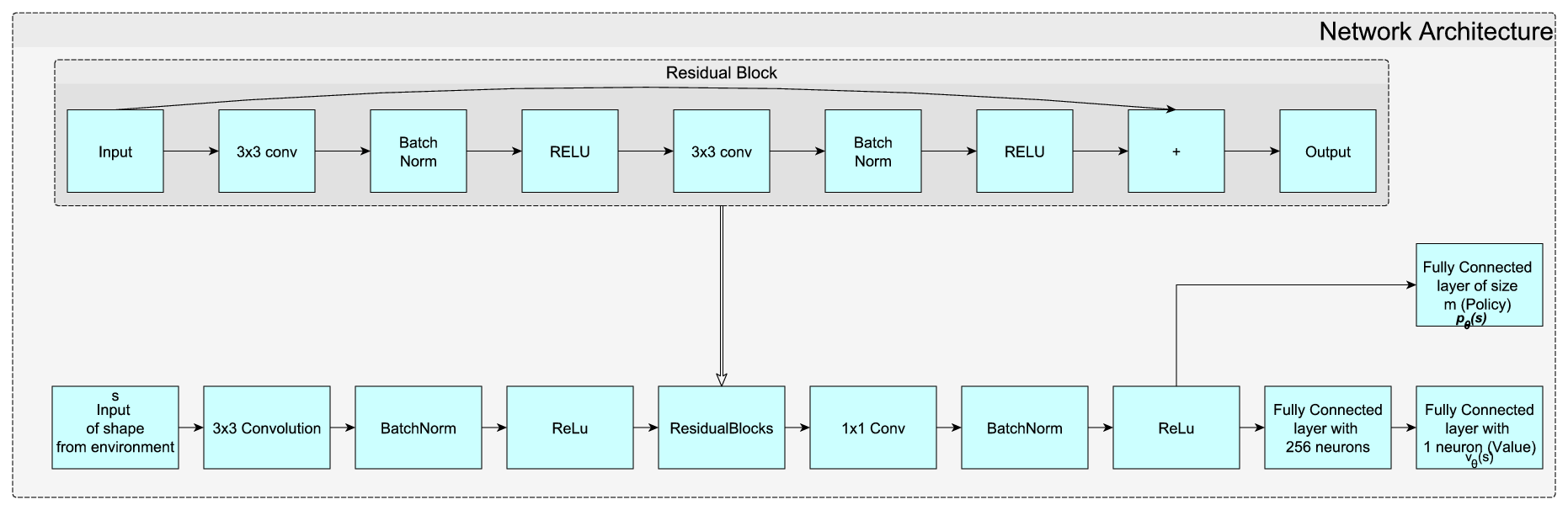}
    \caption{Architecture for the AI players with input {${s}$} from the environment and outputs ${p_{\theta}}{(s)}$ and {$v_{\theta}(s)$}. Network parameters are shown in Tables \ref{table:1} and \ref{table:2}. Each residual block has 3x3 convolution has 512 filters when playing Reversi and 256 filters when playing Racing Kings.}
    \label{fig:architecture}     
    \end{minipage}
\end{figure*}

A deep residual convolutional neural-network is used for both players with parameters as shown in Tables \ref{table:1} and \ref{table:2}. Deep residual convolutional neural-networks have been found to be stable and less prone to overfitting than traditional convolutional neural-networks\cite{he_deep_2015} for large networks. The architecture is shown in Figure \ref{fig:architecture}. 

The input to the neural-network is a state tensor ${s}$ from the game environment. The neural-network has two outputs, a policy vector {${p_{\theta}}{(s)}$}, and a scalar value estimate $v_{\theta}(s)$ for the given {${s}$}.
{${p}_{\theta}(s)$} is an $m$ sized vector, indexed by actions $a$, representing the probability distribution of the best actions to take from {${s}$}. 

\begin{figure}
   \begin{minipage}[b]{\columnwidth}
   \includegraphics[width=\textwidth]{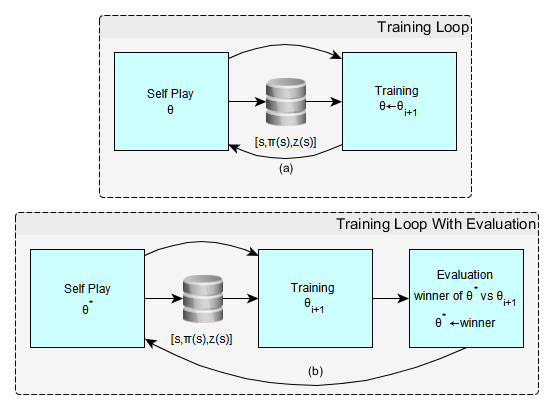}
    \caption{Two training loops used to demonstrate effectiveness of curriculum learning. a. AlphaZero inspired and b. AlphaGo Zero inspired.}
    \label{fig:azloop}     
    \end{minipage}
\end{figure}

\vspace{\baselineskip}\noindent\paragraph{Training Pipeline}
\label{training}
The training pipeline consists of two independent processes; self-play and optimisation. The initial neural-network weights ${\theta}_{i=0}$ are randomised and after each training iteration, $i$, are updated yielding ${\theta}_{i+1}$. The self-play process plays games against itself using the latest weights to generate training examples, filling an \emph{experience buffer}. The optimisation process trains the network using these experiences via batched gradient descent. The process is shown in Figure \ref{fig:azloop}(a). 

Unlike \emph{AlphaZero} where the processes are executed in parallel across multiple systems for maximum performance, we seek to reduce the variability of training by conducting the training process sequentially on a single system.   When the \emph{experience buffer} contains the experiences from a set number of games, self-play stops and optimisation begins. After optimisation has finished all experiences from a percentage of the earliest games are removed from the \emph{experience buffer}, and self-play recommences.

Training experiences are generated during self-play using the latest network weights. Each training experience {${X}$} is comprised of a set {${{X}:=\{{s},{\pi}{(s)},z{(s)}\}}$} where {${s}$} is the tensor representation of the state, {${\pi}{(s)}$} is a probability distribution of most likely actions \emph{(policy)} indexed by $a$ obtained from the MCTS, and $z{(s)}$ is the scalar reward from the perspective of the player for the game's terminal state. During self-play, an experience is saved for every ply\footnote{The term `ply' is used to disambiguate one player's turn which has different meanings in different games. One ply is a player's single action.} in the game, creating an \emph{experience buffer} full of experiences from a number of different self-play games. $z{(s)}=r_{T}{(s)} $ where $r_{T}{(s)}$ is the reward for the terminal state of the game and is $-1$ for a loss, $+1$ for a win, and $-0.5$ for a draw. We use a slightly negative reward for a draw instead of $0$ to discourage the search from settling on a draw, instead preferring exploration of other nodes which are predicted as having a slightly negative value.

During training, the experiences are randomised, and parameters $\theta$ are updated to minimise the difference between $v_{\theta}(s)$ and $z{(s)}$, and maximise similarities between {${p}_{\theta}(s)$} and {${\pi}{(s)}$} using the loss function shown in Equation \ref{eq:loss}. One \emph{training step} is the presentation of a single batch of experiences for training, while an epoch {$t$} is completed when all experiences in the buffer have been utilised. $20$ epochs are conducted for each training iteration $i$. The experiences are stored in the \emph{experience buffer} in the order in which they were created. At the conclusion of each iteration the buffer is partially emptied by removing all experiences from a portion of the oldest games. 

\begin{equation}
    l=(z{(s)}-v_{\theta}(s))^2-{\pi}{(s)}\cdot log({p}_{\theta}(s))+c\cdot ||\theta||^2
    \label{eq:loss}
\end{equation}
where:
\begin{conditions}
z{(s)} & The reward from the game's terminal state (return).\\
c\cdot ||\theta||^2 & L2 weight regularisation. \\
v_{\theta}(s) & Neural-network value inference. \\
{\pi}{(s)} & Policy from the Monte-Carlo Tree Search (MCTS).\\
{p}_{\theta}(s) & Neural-network policy inference.
\end{conditions}

For completeness we also separately conduct an additional experiment using \emph{AlphaGo Zero's} method of adding a third process to evaluate the best weights $\theta^\star$ \cite{silver_mastering_2016}. The primary difference for this experiment is that self-play is conducted with the best weights $\theta^\star$ instead of the current weights, as shown in Figure \ref{fig:azloop}(b).

\vspace{\baselineskip}\noindent\paragraph{Monte-Carlo Tree Search (MCTS)}
\label{MCTS}
The MCTS builds an asymmetric tree with the states ${s}$ as nodes, and actions $a$ as edges. At the conclusion of the search the policy ${\pi}{(s)}$ is generated by calculating the proportion of the number of visits to each edge.

During the tree-search the following variables are stored:
\begin{itemize}
    \item The number of times an action was taken {$N{(s,a)}$}.
    \item The neural-network's estimation of the optimum policy {${p}_{\theta}(s)$} and the network's estimate of the node's value {$v_{\theta}(s)$}.
    \item The value of the node {$V{(s)}$}. 
    \item The average action-value for a particular action from a node {$Q{(s,a)}$}.
\end{itemize}

MCTS is conducted as follows:
\begin{itemize}
\item \textbf {Selection.} The tree is traversed from the root node by calculating the upper confidence bound $U{(s,a)}$ using Equation \ref{eq:UCT} and selecting actions {${a_i=\underset{a}{\mbox{argmax}} (U{(s,a)})}$} until a leaf node is found\cite{silver_mastering_2017}. Note the use of ${p}_{\theta}(s)$ from the neural-network in Equation \ref{eq:UCT}. 
\item \textbf {Expansion.} When a leaf node is found $v_{\theta}(s)$ and ${p}_\theta{(s)}$ are obtained from the neural-network and a node is added to the tree.
\item \textbf {Evaluation.} If the node is terminal the reward $r$ is obtained from the environment for the current player and $V{(s)}:=r$ otherwise $V(s):=v_{\theta}(s)$. Note that there is no Monte Carlo roll-out. 

\item \textbf {Backpropogation.} $Q{(s,a)}$ is updated by the weighted average of the old value and the new value using Equation \ref{eq:wAverage}.

\end{itemize}
\begin{equation}
    U{(s,a)}=Q{(s,a)}+3 \cdot p^{\prime}_{a}{(s)} \frac{ \sqrt{\sum\limits_{j=1}^{m}{N{(s,a_j)}}}}{1+N{(s,a)}}
\label{eq:UCT}
\end{equation}
\begin{equation}
    {p}^{\prime}{(s)} = {p}_{\theta}(s)+{\delta} 
\end{equation}
    \begin{equation}
        Q{(s,a)} \leftarrow\frac{N{(s,a)} \cdot Q{(s,a)}+ V{(s)}}{N{(s,a)} + 1}
    \label{eq:wAverage}
    \end{equation}
where 
\begin{conditions}
Q{(s,a)}& Average action value.\\
{p}_{\theta}(s)& Policy for ${s}$ from neural-network.\\ 
{\delta} & The Dirichlet noise function. \\
{p}^{\prime}{(s)}& ${p}_{\theta}(s)$ with added Dirichlet noise. \\
\sum\limits_{j=1}^{m}{N{(s,a_j)}}& Total visits to parent node.\\
N{(s,a)} & Number of visits to edge ($s,a$).\\
V{(s)} & The estimated value of the node.
\end{conditions}
At the conclusion of the search a probability distribution ${\pi}{(s)}$ is calculated from the proportion of visits to each edge ${\pi}_{a_{j}}{(s)}=\frac{N(s,a_{j})}{\sum{N(s,{A})}}, j=1,..,m$. The tree is retained for reuse in the player's next turn after it is trimmed to the relevant portion based on the opponent's action. 

\vspace{\baselineskip}\noindent\paragraph{Move Selection}

 To ensure move diversity during each self-play game, for a given number of ply as detailed in Tables \ref{table:1} and \ref{table:2}, actions are chosen probabilisticly based on ${\pi}{(s)}$ after the MCTS is conducted. After a given number of ply, actions are then chosen in a greedy manner by selecting the action with the largest probability from ${\pi}{(s)}$. In a competition against a \emph{reference opponent}, moves are always selected greedily.  


\subsubsection{Curriculum Player}
\label{sec:playerwithcurriculum}
The difference between the \emph{baseline player} and the \emph{curriculum player} is that some experiences are excluded from the \emph{experience buffer} during training of the \emph{curriculum player}. A curriculum function $\zeta{(t)}$ is introduced which indicates the percentage of experiences to be retained for a given game, depending on the number of epochs $t$ which have occurred. When storing a game's experiences the first $(1-\zeta{(t)})\%$ of experiences are excluded from a game. This can be achieved by either trimming the game's experiences or not generating the experience in the first place. The \emph{baseline player} has an equivalent function {$\zeta{(t)}=1$} for all $t$; that is retaining $100\%$ of a game's experiences. The curriculum used for Racing Kings is shown in Equation \ref{eq:RKStep}, while the curriculum used for Reversi is shown in Equation \ref{eq:revStep}. 


We exploit the fact that the \emph{curriculum player} excludes some early-game experiences during training by not conducting a tree-search for actions which will result in an experience that is going to be excluded. Instead of naively conducting a full search and discarding the experiences, we randomly choose the actions for the estimated number that would have been discarded. We manage the number of random actions by maintaining the average ply, $av$, in a game, and playing $(1-\zeta{(t)})\%\cdot av$ of the actions randomly. After these random actions, we then use the full MCTS to choose the remaining actions to play the game. If a terminal state is found during random play then the game is rolled back $\zeta{(t)}\%$ ply and MCTS is used for the remaining actions.


\begin{equation}
\zeta_{\text{\tiny racingkings}}(t)=
  \begin{cases}
      0.1 & t= 0 \\
      0.33 & 0< t< 100 \\
      0.5 & 100\leq t<500\\
      0.66 & 500\leq t<800\\
      0.8 & 800\leq t<1000\\
      1.0 & 1000\leq t\\
   \end{cases}
   \label{eq:RKStep}
\end{equation}

\begin{equation}
\zeta_{\text{\tiny reversi}}(t)=
  \begin{cases}
      0.25 & t= 0 \\
      0.5 & 0< t< 100 \\
      0.75 & 100\leq t<500\\
      1.0 & 500\leq t\\
      
   \end{cases}
   \label{eq:revStep}
\end{equation}

\subsubsection{AlphaGo Zero Inspired Player}
\label{agzinspired}
The \emph{AlphaGo Zero} inspired player is very similar to the player explained above (see Section \ref{sec:playerwithcurriculum}), with the exception of an evaluation step in the training loop. The evaluation step plays a two player competition between the current best player with weights ${\theta}^\star$ and a challenger using the latest weights ${\theta}_{i}$. The competition is stopped when the lower bound of the $95\%$ percentile Wilson confidence score with continuity correction\cite{reed_better_2007} is above $0.5$, or the upper bound is below $0.5$ allowing a competition winner to be declared. 
The competition is also stopped when the difference between the upper and lower confidence interval is less than $0.1$, in which case no replacement is conducted. If the challenger is declared the winner of the competition, then it's weights become the best weights and are used for subsequent self-play until they are replaced after a future evaluation competition. Although this method for stopping does not provide exactly $95\%$ confidence\cite{frey_fixed-width_2010}, it provides sufficient precision for determining which weights to use to create self-play training examples.       

\subsubsection{Reference Opponent}
The \emph{reference opponent} provides a fixed performance opponent for testing the quality of the AI players. Stockfish-multivariant \cite{dugovic_multi-variant_2019} is used as the \emph{reference opponent} for Racing Kings; whilst an MCTS player using 200 simulations \cite{browne_survey_2012} is used for Reversi.

\section{Experiments}
\label{sec:eval}

Two AI players are trained: one using the proposed curriculum learning approach \emph{(curriculum player)}, the other without \emph{(baseline player)}. A competition is periodically conducted during training between the AI players and a \emph{reference opponent}, and the players' respective win ratios $y_{\theta_{i}}$ are recorded. A competition consists of a minimum of 30 games for each randomly selected $\theta_{i}$ to obtain the win ratio. The experiment is conducted in full three times and the results are combined. The moving average of the win ratio ${y}_{\theta_{i}}$ from all three experiments is plotted against time, steps and epochs. A \emph{training step} is completed after the presentation of one batch of experiences and a \emph{training epoch} is completed after all experiences have been presented once. As training begins after a set number of games, the number of steps for each epoch varies from experiment to experiment, likewise the time taken to play a game is also completely unique making the time, epochs and steps independent from experiment to experiment. As such, when combining data from multiple experiments the three plots may have different appearances.




Whilst ${y}_{\theta_{i}}$ vs time is the measure which we are primarily interested in, measuring against steps and epochs are also informative.  We mitigate the potential differences which might arise from differing system loads by training both AI players simultaneously on one dual-GPU system. 

\subsection{Results}
\begin{figure}[htp]
  \centering 
    \minipage{0.48\textwidth}
      \centering 
      \includegraphics[width=1\textwidth]{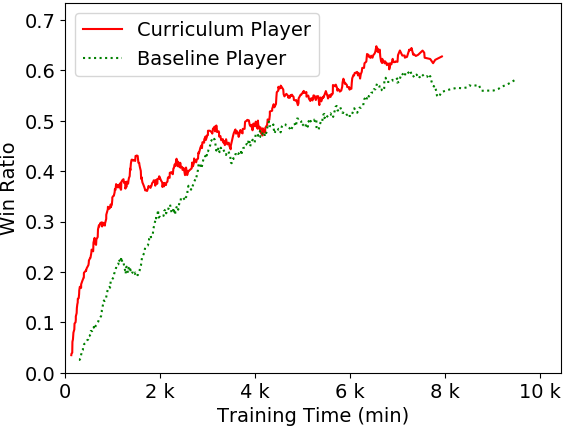} 
            \centering 
    (a)\endminipage\hfill
    \minipage{0.48\textwidth}
      \includegraphics[width=1\textwidth]{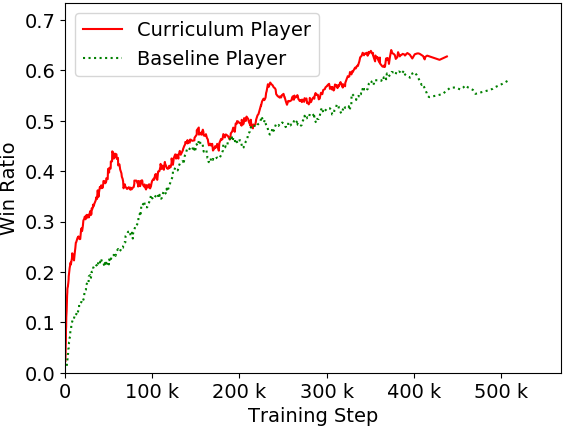}
    \centering 
    (b)\endminipage\hfill
    \minipage{0.48\textwidth}
    \includegraphics[width=1\textwidth]{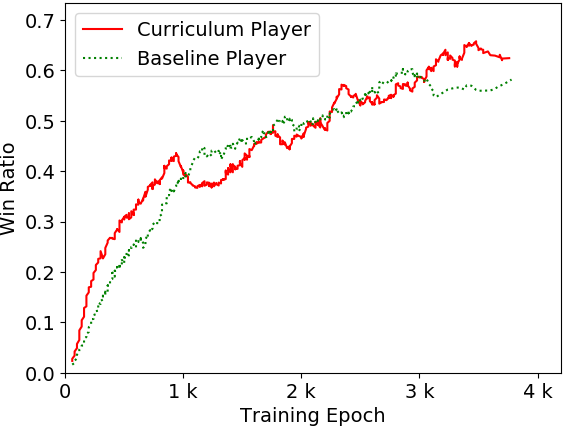}
     \centering 
    (c)\endminipage\hfill

    \caption{Win ratio for AlphaZero inspired player vs Stockfish level 2 playing Racing Kings with and without training curriculum from Equation \ref{eq:RKStep}. Note that improvement is seen in both the time and the steps figures, indicating that the performance improvement is more than the time saved by conducting random moves during self-play. Note the unlearning which occurs around 1000 epochs for the player with a curriculum.  This plot is the 20 point moving average of 3 independent training runs.
    }

    \label{fig:AZ_RK}
\end{figure}

\begin{figure}[htp]
  \centering 
    \minipage{0.48\textwidth}
    \includegraphics[width=1\textwidth]{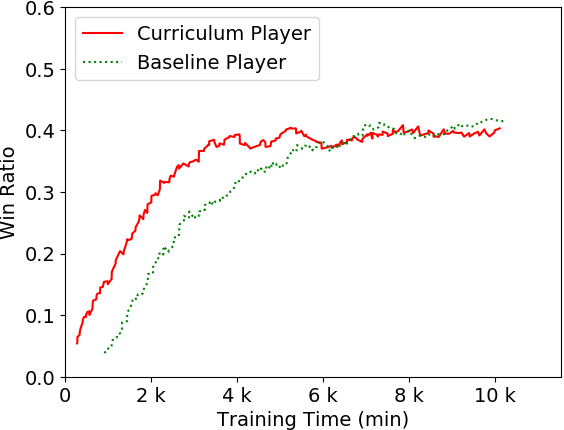}    
    \centering 
    (a)\endminipage\hfill
    \minipage{0.48\textwidth}
    \includegraphics[width=1\textwidth]{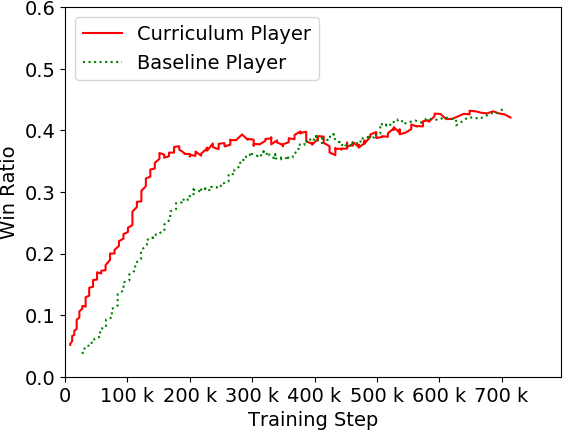}
    \centering 
    (b)\endminipage\hfill
    \minipage{0.48\textwidth}
    \includegraphics[width=1\textwidth]{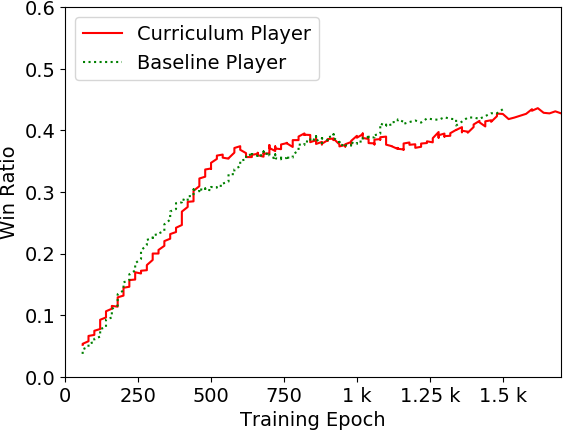}
    \centering 
    (c)\endminipage\hfill
    
    \caption{Win ratio for AlphaZero inspired player vs MCTS with 200 simulations playing Reversi with and without the training curriculum from Equation \ref{eq:revStep}. Note that the epoch win ratio of both players is similar despite the player with curriculum learning having less training examples per epoch due to the dropped experiences. This plot is the 20 point moving average of 3 independent training runs.}
    \label{fig:AZ_Rev}
\end{figure}

\begin{figure}[htp]
  \centering 
    \minipage{0.48\textwidth}
    \includegraphics[width=1\textwidth]{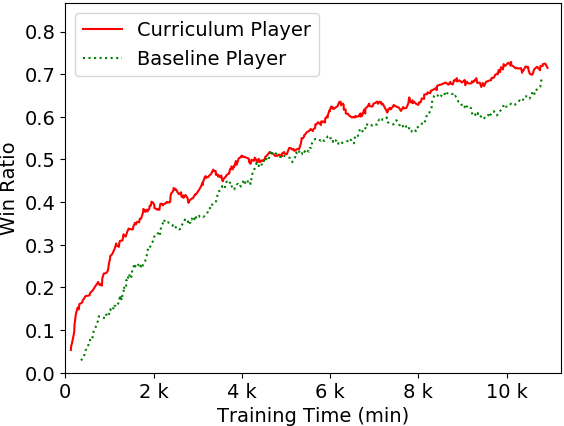}
     \centering 
    (a)\endminipage\hfill
    \minipage{0.48\textwidth}
    \includegraphics[width=1\textwidth]{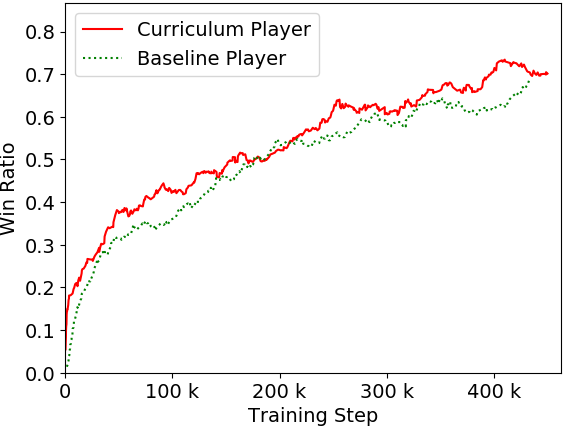}
     \centering 
    (b)    \endminipage\hfill
    \minipage{0.48\textwidth}
    \includegraphics[width=1\textwidth]{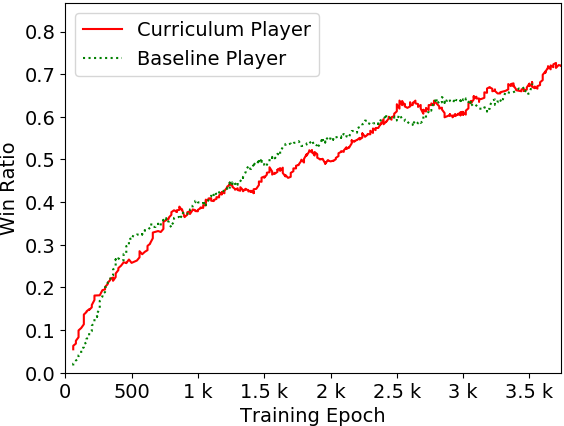}
         \centering 
    (c)    \endminipage\hfill

        \caption{Win ratio for AlphaGoZero inspired player vs Stockfish level 2 playing Racing Kings with and without training curriculum from Equation \ref{eq:RKStep}. This plot is the 20 point moving average of 3 independent training runs. \newline \newline }
    \label{fig:AGZ_RK}
\end{figure}

Figure \ref{fig:AZ_RK} shows the win ratio of the \emph{AlphaZero} inspired players in a Racing Kings competition against the \emph{Stockfish} opponent; Figure \ref{fig:AZ_Rev} shows the win ratio of the \emph{AlphaZero} inspired players playing Reversi against the MCTS opponent; and Figure \ref{fig:AGZ_RK} shows the win ratio whilst playing Racing Kings against the Stockfish opponent but using an \emph{AlphaGoZero} inspired player with the added evaluation step. 


The win ratio of the player using the \emph{end-game-first} training curriculum exceeds the \emph{baseline player} during the early stages of training in all cases when measured against time. This was also observed across multiple training runs with differing network parameters and differing curricula.   


\subsection{Discussion}
Our results indicate that a player trained using an \emph{end-game-first} curriculum learns faster than a player with no curriculum during the early training periods. The improvement over time can, in part, be attributed to the increased speed of self-play moves being chosen randomly by the \emph{curriculum player} instead of conducting a full search; however the win ratio improvement is also observed when compared over \emph{training steps} indicating that a more subtle benefit is obtained. The win ratio of the two players when compared against \emph{training epoch} shows similar performance for the two players.   

In this section we compare and contrast the performance of the \emph{curriculum player} and the \emph{baseline player} against the \emph{reference opponents} and separately discuss the results with respect to time (Subsection \ref{subsec:time}), training steps (Subsection \ref{subsec:steps}) and epochs (Subsection \ref{subsec:epochs}).  

\subsubsection{Win ratio vs Time comparison}
\label{subsec:time}
Subfigure (a) of Figures \ref{fig:AZ_RK}, \ref{fig:AZ_Rev} and \ref{fig:AGZ_RK} shows the win ratio of the AI players vs Time. The \emph{curriculum player} does not retain experiences generated early in a game during the early training periods. This is achieved by selecting these moves randomly instead of using a naive approach of conducting a full search, and dropping the early-game experience as explained in Section \ref{sec:playerwithcurriculum}. The time saved as a result of randomly selecting moves instead of conducting a tree-search can be significant, however by using random move selection no experience is added to the \emph{experience buffer} meaning that a randomly selected move contributes in no way to the training of the player. The curriculum needs to balance the time saved by playing random moves with the reduction in generating training experiences. 

Whilst the \emph{curriculum player} leads in win ratio compared to the \emph{baseline player} during the early time periods, the win ratios of both players converge in all experiments. For a given neural network configuration there is some expected maximum performance threshold against a fixed opponent; in the ideal case this would be a 100\% win ratio but if the network is inadequate it may be less. Although we expect that the win ratios of the two players would converge eventually, it appears that convergence occurs prior to the maximum performance threshold as shown in Figure \ref{fig:AZ_RK}(a) near $4000$ minutes and Figure \ref{fig:AGZ_RK}(a) near $5000$ minutes; while Figure \ref{fig:AZ_Rev}(a) shows convergence near $6000$ minutes at what appears to be the networks maximum performance threshold. For an optimal curriculum, the convergence of the two players would be expected to occur only at the maximum performance threshold.

\subsubsection{Win ratio vs Steps comparison}
\label{subsec:steps}
Subfigure (b) in Figures \ref{fig:AZ_RK}, \ref{fig:AZ_Rev} and \ref{fig:AGZ_RK} shows the win ratio of the AI players vs the number of training steps. Although the win ratio improvement over time for the \emph{curriculum player} can be attributed in part to the use of random move selection, a win ratio improvement is also observed when measured against \emph{training steps}. A training step is when one batch of experiences is presented to the optimise module, making a training step time independent; i.e. how long it takes to create an experience is not a factor for training steps. 

Given that the win ratio for the \emph{curriculum player} outperforms the \emph{baseline player} when measured over \emph{training steps}, we argue that there is a gain which relates directly to the net quality of the training experiences. Consider, for example, the \emph{baseline player's} very first move decision in the very first self-play game for a newly initialised network. With a small number of MCTS simulations relative to the branching factor of the game, the first AI player's decision will not build a tree of sufficient depth to reach a terminal node, meaning that the decision will be derived solely from the untrained network weights - recall that the neural-network is used for non-terminal value estimates. The resulting policy ${\pi}$ which is stored in the \emph{experience buffer} has no relevance to the game environment as it is solely reflective of the untrained network. We posit that excluding these uninformed experiences results in a net improvement in the quality of examples in the training buffer. Later in that first game, terminal states will eventually be explored, over-ruling the inaccurate network estimates, and the resulting policy will become reflective of the actual game environment - these are experiences which should be retained. As the training progresses the network is able to make more accurate predictions further and further from a terminal state creating a \emph{visibility horizon} which becomes larger after each epoch. The optimum curriculum would match the change of the visibility horizon.      

\subsubsection{Win ratio vs Epoch comparison}
\label{subsec:epochs}
Subfigure (c) in Figures \ref{fig:AZ_RK}, \ref{fig:AZ_Rev} and \ref{fig:AGZ_RK} shows the win ratio of the players vs the number of training epochs. Recall that an epoch is when all experiences in the \emph{experience buffer} have been presented to the optimise module. Since training is only conducted when the \emph{experience buffer} has sufficient games, an epoch is directly proportional to the number of self-play games played, but is independent of the number of experiences in the buffer and the time it takes to play a game. When applying the curriculum, fewer experiences per game are stored during the early training periods compared to the baseline player, meaning that the \emph{curriculum player} is trained with fewer training experiences during early epochs. 

The plots of the win ratio vs epoch shows the \emph{curriculum player} outperforming the \emph{baseline player} for the early epochs in Figure \ref{fig:AZ_RK}(c) with the two win ratios converging rapidly; while Figures  and \ref{fig:AZ_Rev}(c) and \ref{fig:AGZ_RK}(c) show the two players performing similarly. The similarity of the results when measured against epochs shows that despite the \emph{curriculum player} excluding experiences, no useful information is lost in doing so. In recognising that useful information is not excluded during these early epochs it supports our view that the net quality of the data in the \emph{experience buffer} is improved by applying the specified \emph{end-game-first} curriculum.

It is expected that due to a combination of the game mechanics and the order in which a network learns there may be some \emph{learning resistance}\footnote{To our knowledge, the term \emph{learning resistance} is not defined in relation to machine learning. We define it to mean a short term resistance to network improvement.} which could result in plateaus in the win ratio plot, allowing the trailing player to catch up temporarily. We expect a sub-optimal curriculum to result in additional learning resistance or in the extreme case \emph{learning loss} which would predominantly be observed immediately following a change in the curriculum value. For Reversi the final curriculum increment from Equation \ref{eq:revStep} occurs after 500 Epochs and Figure \ref{fig:AZ_Rev}(c) shows a training plateau shortly after this change, albeit at the network's maximum learning limit. Likewise in Figure \ref{fig:AZ_RK}(c) a learning loss is observed at an average of $900$ epochs shortly after the curriculum has changed to $80\%$ as shown in Equation \ref{eq:RKStep}, although of the three experiments that comprise the data for this plot two of them have a learning loss around $800$ epochs and the other at $1000$ epochs - the final step in the curriculum. Figure \ref{fig:AZ_RK}(c) appears to indicate that on average the learning loss is caused by the final steps of the curriculum, however the learning loss was not observed in all training runs. The presence of this loss in the average of training runs, but the absence from some individual training runs highlights the importance of the order in which learning occurs and its impact on the effectiveness of the curriculum.




\subsubsection{Curriculum Considerations}
Although it is expected that the \emph{baseline player's} performance would converge with the \emph{curriculum player's} performance, ideally this would occur near the maximum win ratio or at some training plateau. The fact that the player's win ratios converge before a clear training plateau has been reached indicates that the curriculum is sub-optimal, and given that the curriculum implemented is semi-arbitrary this is expected. While curriculum learning is shown to be beneficial during the early stages of training, the gain can be lost if the curriculum changes too slowly or too abruptly. 

When designing a curriculum, consideration needs to be given to the speed at which the curriculum changes. At one extreme is the current practice where the full game is attempted to be learnt at all times, i.e. the curriculum is too fast by immediately attempting to learn from $100\%$ of a game at epoch 0. At the other extreme is where the curriculum is too slow, which can result in the network overfitting to a small portion of the game space or discarding examples which contain useful information. We argue that each training run for each game could have its own optimum curriculum profile, due to the different training examples which are generated.

\section{Conclusion}
\label{sec:conc}

The rate at which an AI player learns when using a combined neural network-MCTS architecture can be improved by using an \emph{end-game-first} training curriculum. Although the hand-crafted curricula used in this study are not optimal, a fixed curriculum is not likely to be optimal at all times as the order and the composition of the experiences are themselves a factor. 
The following considerations are required for an optimal curriculum;
\begin{itemize}
\item Balancing the time saved by random moves with the loss of training experiences.
\item Minimising training plateaus that are not related to game complexity.
\item The curriculum profile changes are not too fast as to include uninformed examples.
\item The curriculum profile changes are not too slow as to cause the network to overfit to a smaller portion of the environment space.
\item The curriculum profile changes are not too slow as to result in discarding examples that are sufficiently informed.

\end{itemize}
To address these requirements, the curriculum profile should relate to the visibility horizon of the tree search, not just the number of training iterations. Our future research will explore how a curriculum can be automated based on the visibility horizon of the player's search.
%
%


%



\begin{table*}
\begin{tabular}{p{0.3\hsize}|p{0.1\hsize}|p{0.5\hsize}}%
\bfseries Parameter & \bfseries Value & \bfseries Comment
\begin{filecontents*}{files/rk_params.csv}
Parameter,Value,Comment
Resnet Blocks, 10, Number of Residual Network Blocks
Batch Size, 350, Training Batch size
Value function size, 256, Neurons before $v_{\theta(s)}$
CNN Filters, 512, How many filters are used 
Simulations per action,200, How many nodes are added to the tree during the MCTS tree search
Drop draw chance,0.5, If a game is a draw then it has this chance of being dropped. Used to mitigate if a game has a lot of draws from having too many experiences which are draws
Min Buffer Size, 350; 500; 750; 1000; 1500; 2000, Minimum number of games in the training buffer to stop self-play and commence training. For Chess the number of required games was incremented to allow training to commence earlier
Experience replacement, 0.1, Proportion of experiences removed from training buffer after self-play
Number probabilistic turns, 30, Number of players actions which are selected probabilistically from the policy before switching to selecting moves greedily
\end{filecontents*}
\csvreader[head to column names]{files/rk_params.csv}{}
{\\\hline\Parameter\ &\Value\ &\Comment}

\end{tabular}
\caption{Parameters for Racing Kings player}
\label{table:1}

\end{table*}
\begin{table*}
\begin{tabular}{p{0.3\hsize}|p{0.1\hsize}|p{0.5\hsize}}%
\bfseries Parameter & \bfseries Value & \bfseries Comment
\begin{filecontents*}{files/rev_params.csv}
Parameter,Value,Comment
Resnet Blocks, 10, Number of Residual Network Blocks
Batch Size, 350, Training Batch size
Value function size, 256, Neurons before $v_{\theta(s)}$
CNN Filters, 512, How many filters are used 
Simulations per action,200, How many nodes are added to the tree during the MCTS tree search
Drop draw chance,0.1, If a game is a draw then it has this chance of being dropped. Used to mitigate if a game has a lot of draws from having too many experiences which are draws
Min Buffer Size, 2000, Minimum number of games in the training buffer to stop self-play and commence training
Experience replacement, 0.1, Proportion of experiences removed from training buffer after self-play
Number probabilistic turns, 16, Number of players actions which are selected probabilistically from the policy before switching to selecting moves greedily
\end{filecontents*}
\csvreader[head to column names]{files/rev_params.csv}{}
{\\\hline\Parameter\ &\Value\ &\Comment}

\end{tabular}
\caption{Parameters for Reversi player}
\label{table:2}

\end{table*}

\ifCLASSOPTIONcompsoc
  \section*{Acknowledgments}
\else
  \section*{Acknowledgment}
\fi

Computational resources and services used in this work were provided by the HPC and Research Support Group, Queensland University of Technology, Brisbane, Australia.

\ifCLASSOPTIONcaptionsoff
  \newpage
\fi



%
\bibliography{IEEEabrv,0.0Main.bib}
\bibliographystyle{IEEEtran}

%
\begin{IEEEbiography}
[{\includegraphics[width=1in,height=1.25in,clip,keepaspectratio]{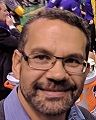}}]
{Joseph West}
completed Bachelor of Engineering and Bachelor of Computer Science degrees from the University of New South Wales - Australian Defence Force Academy, Canberra, ACT, Australia in 2005 and a Masters in Engineering Science in 2009. He is currently researching General Game Playing using Reinforcement Learning at Queensland University of Technology for his Ph.D.   
\end{IEEEbiography}
\begin{IEEEbiography}
[{\includegraphics[width=1in,height=1.25in,clip,keepaspectratio]{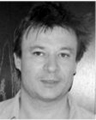}}]
{Frederic Maire}
 received the Ph.D. degree in discrete mathematics from the Universite Pierre et Marie Curie, Paris 6, France, in 1993, after completing M.Sc. degrees in pure mathematics and computer science engineering in 1989.
Currently, he is a senior lecturer at the School of Electrical Engineering and Computer Science, Queensland University of Technology, Brisbane, Australia. 
His research interests include computer vision and robotics.
Dr. Maire is a member of the IEEE.
\end{IEEEbiography}
\begin{IEEEbiography}
[{\includegraphics[width=1in,height=1.25in,clip,keepaspectratio]{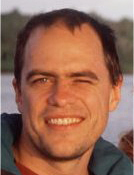}}]
{Cameron Browne} is an Associate Professor at Maastricht University's Department of Data Science and Knowledge Engineering (DKE), where he is running the \euro 2m ERC-funded {\it Digital Ludeme Project} over the next five years. 
He received the Ph.D. degree from QUT in 2009, 
producing the world's first published computer-generated games. 
Dr Browne is the author of the books {\it Hex Strategy}, {\it Connection Games} and {\it Evolutionary Game Design}, which won the 2012 GECCO �Humies� award for human-competitive results in evolutionary computation. 
He is a Section Editor of the {\it IEEE Transactions on Games} and the {\it International Computer Games Association (ICGA)} journal, and is the founder and Editor-in-Chief of the {\it Game \& Puzzle Design} journal.
\end{IEEEbiography}
\begin{IEEEbiography}
[{\includegraphics[width=1in,height=1.25in,clip,keepaspectratio]{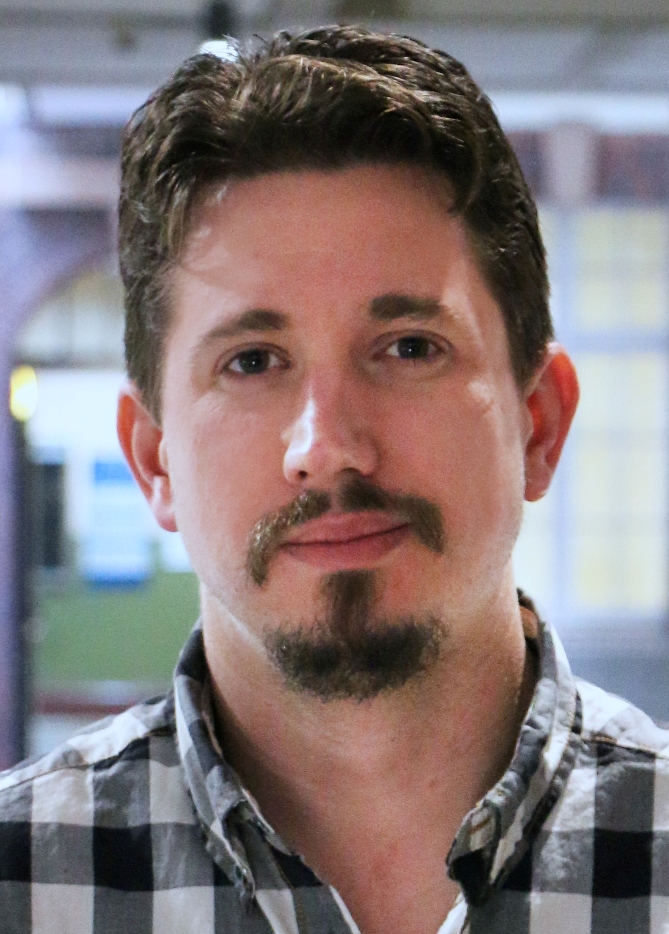}}]
{Simon Denman} received a BEng (Electrical), BIT, and PhD in the area of object tracking from the Queensland University of Technology (QUT) in Brisbane, Australia. He is currently a Senior Lecturer with the School of Electrical Engineering and Computer Science at the Queensland University of Technology. His active areas of research include computer vision and spatio-temporal machine learning. Dr. Denman is a member of the IEEE.
\end{IEEEbiography}









\end{document}